\documentclass{article}

\PassOptionsToPackage{numbers, compress}{natbib}
\usepackage[preprint]{neurips_2024}

\usepackage[utf8]{inputenc} % allow utf-8 input
\usepackage[T1]{fontenc}    % use 8-bit T1 fonts
\usepackage{hyperref}       % hyperlinks
\usepackage{url}            % simple URL typesetting
\usepackage{booktabs}       % professional-quality tables
\usepackage{amsfonts}       % blackboard math symbols
\usepackage{nicefrac}       % compact symbols for 1/2, etc.
\usepackage{microtype}      % microtypography
\usepackage{xcolor}         % colors

\usepackage{graphicx}

\title{LinguaLinker: Audio-Driven Portraits Animation with Implicit Facial Control Enhancement}

\author{
  Rui Zhang\thanks{Equal contribution} \quad Yixiao Fang$^{\ast}$ \quad Zhengnan Lu \quad Pei Cheng \quad Zebiao Huang \quad Bin Fu \vspace{3pt}\\
  \normalsize{Tencent} \\
  \normalsize{\{rainarzhang, yixiaofang, zanelu, peicheng, zebiaohuang, brianfu\}@tencent.com} \vspace{3pt}\\
  \url{https://tencentqqgylab.github.io/LinguaLinker}
}

\begin{document}
\maketitle
\begin{abstract}
  This study delves into the intricacies of synchronizing facial dynamics with multilingual audio inputs, focusing on the creation of visually compelling, time-synchronized animations through diffusion-based techniques. Diverging from traditional parametric models for facial animation, our approach, termed \textbf{LinguaLinker}, adopts a holistic diffusion-based framework that integrates audio-driven visual synthesis to enhance the synergy between auditory stimuli and visual responses. We process audio features separately and derive the corresponding control gates, which implicitly govern the movements in the mouth, eyes, and head, irrespective of the portrait's origin. The advanced audio-driven visual synthesis mechanism provides nuanced control but keeps the compatibility of output video and input audio, allowing for a more tailored and effective portrayal of distinct personas across different languages. The significant improvements in the fidelity of animated portraits, the accuracy of lip-syncing, and the appropriate motion variations achieved by our method render it a versatile tool for animating any portrait in any language.
\end{abstract}
\begin{figure}[h]
  \flushleft
  \includegraphics[width=0.98\linewidth]{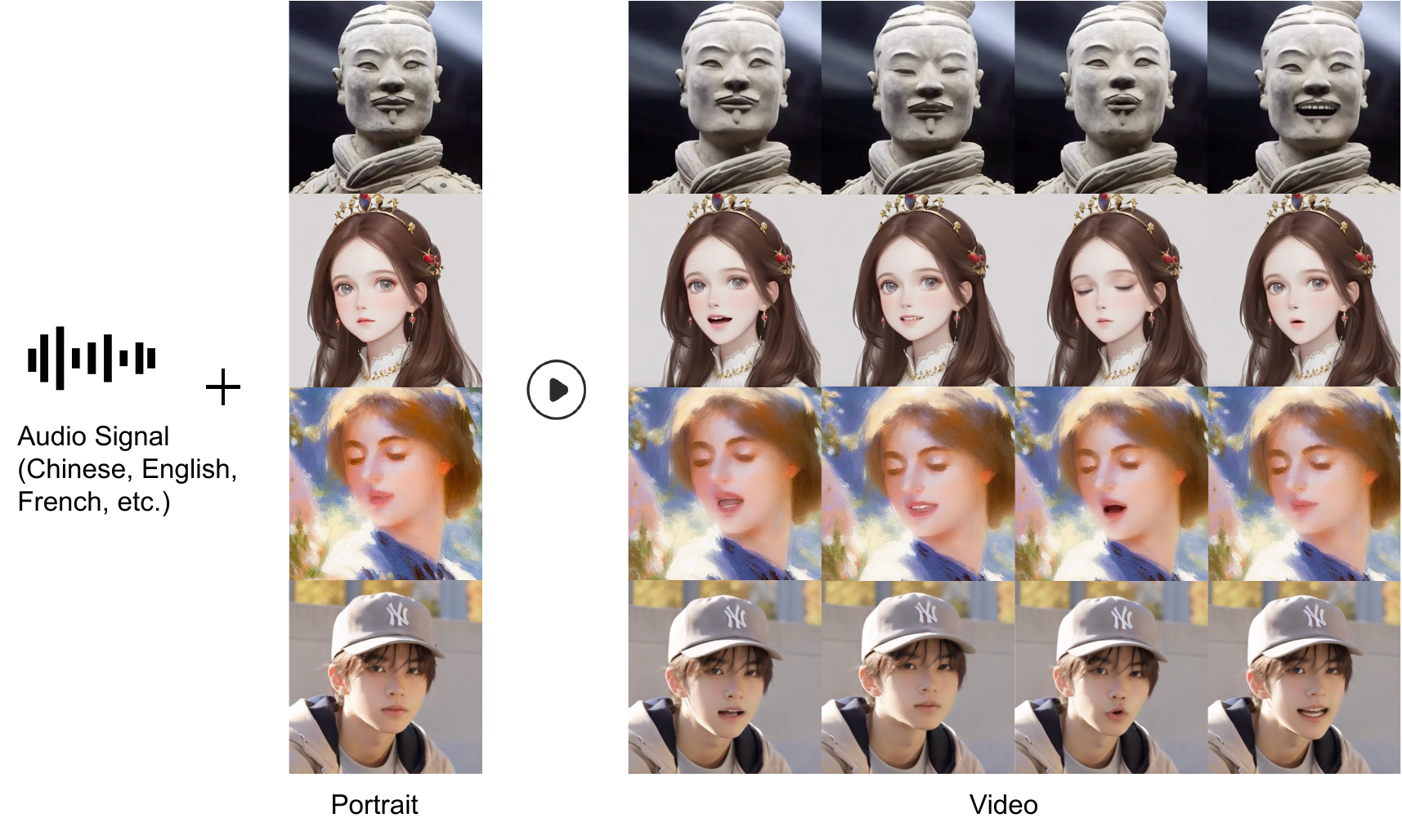}
  \caption{The proposed method LinguaLinker accepts multilingual audio signals and arbitrary portrait images and generates the corresponding video output. }
  \label{fig:teaser}
\end{figure}

\section{Introduction}

Recent advancements in image or video generation have experienced explosive growth, including methods based on Generative Adversarial Networks (GANs)~\cite{goodfellow2014generativeadversarialnetworks} as well as those utilizing Diffusion~\cite{rombach2022highresolutionimagesynthesislatent, ho2020denoisingdiffusionprobabilisticmodels, song2022denoisingdiffusionimplicitmodels} techniques. In the human-related animation field, especially portrait animation as a sub-task of video generation, these techniques also lay the foundation for generating coherent and vivid talking head videos based on the given audio or video inputs and reference portrait images, achieving accurate face restoration and lip synchronization. Thanks to the advancements in these technologies, related applications have also been able to develop rapidly.

However, from a comprehensive perspective, several issues still pose significant challenges to existing models and urgently need to be addressed. For video-based~\cite{xie2024xportraitexpressiveportraitanimation} portrait animation generation, there exists the issue of misalignment between the reference video and the reference portrait, which often results in the generated video failing to accurately reproduce the input reference portrait. This misalignment can lead to discrepancies between the features of the reconstructed result and the reference image. Although methods based on facial landmarks~\cite{wei2024aniportrait, wang2024V-Express} or 3D models~\cite{sun2023vividtalk, zhang2023sadtalkerlearningrealistic3d, ma2023dreamtalk}, abstract facial features, they tend to lose a significant amount of facial detail and do not completely resolve the facial mismatch problem. Additionally, the issue of inter-frame flickering in video generation continues to be a persistent challenge for researchers. Meanwhile, another research direction is how to enable models to generate corresponding video results for audio inputs in multiple languages.

To address some of these issues, we propose an end-to-end audio-driven portrait animation approach \textbf{LinguaLinker}, based on diffusion models, which is capable of generating high-fidelity and lip-synchronized animated portrait videos and supports multilingual audio inputs.

We have meticulously designed the combination process of latents and audio features. First, leveraging latents with audio features and the timestep information of the denoising stage as inputs of Adaptive Layer Normalization(AdaLN)~\cite{Peebles2022DiT}, we obtain the corresponding gate states. Next, we project these gate states to calculate the corresponding gate offsets based on different regional masks. Finally, by adding the offsets and regional masks, we obtain the corresponding region-specific gates. As a clip of the audio has its own emotion, we could use such corresponding mask implicitly to define the range and the hyper-parameter to control the frequency of movements instead of artificial control of facial expressions, showcasing advancements in lip synchronization and the compatibility of the input audio and the generated video. Furthermore, our approach is able to alleviate the flickering issue in video generation, which creates more coherent and higher-fidelity portrait animation. The details of the architecture are explained in section \ref{sec:arch}. 

We also construct a wide and diverse audio-video dataset for model training, including speeches, television series, singing clips, etc. To ensure the dataset quality, we build a rigorous data-cleaning pipeline to remove unqualified samples. After our pipeline of data filtering and curation, the remaining portion constitutes around 114 hours of high-quality audio-video synchronized video clips. The concrete stages of the pipeline are illustrated in section \ref{sec:data}.

\section{Related Work}

\textbf{Diffusion-Based Video Generation.} Diffusion-based models have significantly advanced multimedia tasks, from text-to-image generation to complex video synthesis and 3D digital constructs. Notable advancements include Video Diffusion Models (VDM)~\cite{ho2022videodiffusionmodels} and ImagenVideo~\cite{ho2022imagenvideohighdefinition}, which use space-time factorized U-Nets and cascaded diffusion for high-definition video outputs. Make-A-Video~\cite{singer2022makeavideotexttovideogenerationtextvideo} and ModelscopeT2V~\cite{wang2023modelscopetexttovideotechnicalreport} adopt the 3D-UNet integrating temporal layers to model spatio-temporal dependencies. AnimateDiff~\cite{guo2023animatediff} and VideoCrafter~\cite{chen2023videocrafter1, chen2024videocrafter2} contribute to text-to-video generation by extending from the text-to-image model, making the control model framework originally implemented in the T2I framework also universal in the T2V framework. These models also excel in human-centered image animation, allowing additional control over appearance and motion. The integration of UNet with Transformer-based designs enhances text-conditioned video generation and realistic animated portraits, demonstrating diffusion models' versatility and adaptability in computational creativity.

\textbf{Image-Based Animation.} Image-based animation techniques, as discussed in various studies~\cite{wang2024unianimate, ling2024motionclone, mimicmotion2024}, focus on creating dynamic images or video sequences from static input images. Recent advancements have seen the adoption of diffusion models due to their high-quality output and robust control mechanisms in animating human images. For instance, MagicPose~\cite{chang2024magicposerealistichumanposes}, Animate Anyone~\cite{hu2023animateanyone} and MagicAnimate~\cite{xu2023magicanimate} highlight the effectiveness of the integration of reference image features into the self-attention blocks of Latent Diffusion Model (LDM) UNets to preserve the appearance context. And these approaches, exemplified by the innovative ControlNet~\cite{zhang2023adding} or a simplified control guider, further enhance the LDM framework by introducing controllable image generation capabilities, leveraging additional structural conditions derived from various signals like landmarks, segmentations, and dense poses. This highlights the diffusion model's adaptability and resilience in addressing a wide array of generation challenges. Furthermore, ongoing research is delving into the utilization of diffusion models for talking head animation, yielding promising results that validate the approach's potential in crafting authentic video sequences. These outcomes reinforce the continued significance and investigational merit of this approach.

\begin{figure}[t]
  \centering
  \includegraphics[width=1\linewidth]{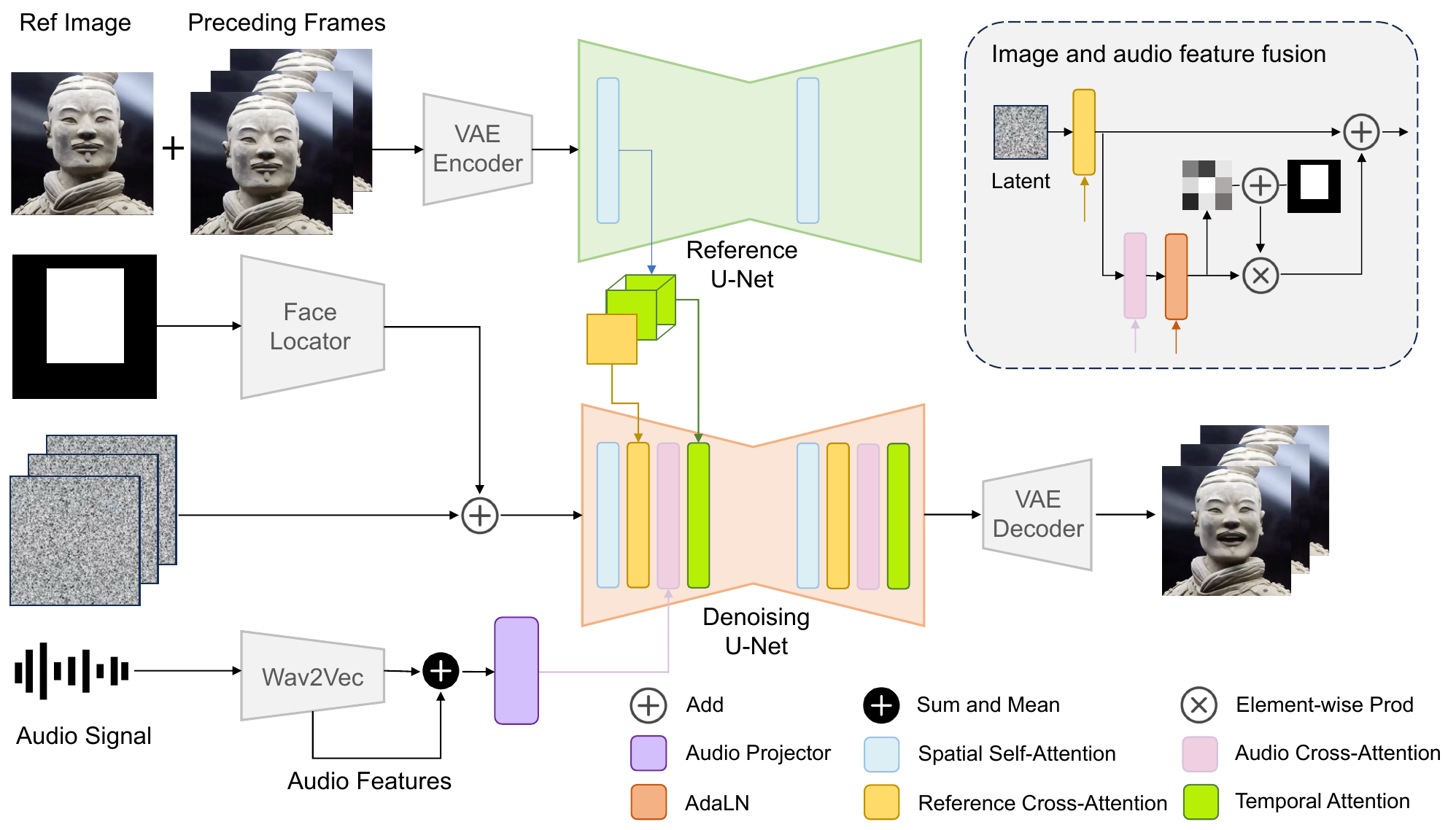}
  \caption{The overview of architecture.}
  \label{fig:arch}
\end{figure}

\textbf{Audio-Driven Portrait Animation.} Audio-driven talking head generation techniques can be broadly classified into two primary methodologies: generating talking head video w/wo intermediate representation. The utilization of an intermediate representation enables the direct or indirect application of an additional visible control signal to inform the video generation process. For example, Vividtalk~\cite{sun2023vividtalk} advocates for the synthesis of head motion and facial expressions, which are subsequently employed to construct a 3D facial mesh. This mesh functions as an intermediate representation to steer the generation of final video frames. Similarly, Sadtalker~\cite{zhang2023sadtalkerlearningrealistic3d} adopts a 3D Morphable Model (3DMM) as an intermediate representation for the production of talking head video. Additionally, Dreamtalk~\cite{ma2023dreamtalk} integrates diffusion models to generate coefficients for the 3DMM, further enhancing the control of the generated video content. Current work such as AniPortrait~\cite{wei2024aniportrait}, also generates the talking head video by extracting the 3D facial mesh and head pose from the audio first, and subsequently project these two elements into 2D keypoints. However, a recurring challenge among these techniques is the restricted ability of the 3D mesh to capture nuanced details, thereby limiting the overall dynamic range and authenticity of the synthesized video sequences. Compared to the methods using intermediate representations, the method without intermediate control signals to drive audio-driven video generation shows higher naturalness and consistent identity preservation with the original image. EMO~\cite{tian2024emo} utilizes a direct audio-to-video synthesis approach to generate expressive portrait videos with an audio2video diffusion model under weak conditions, bypassing the need for intermediate 3D models or facial landmarks.  Hallo~\cite{xu2024hallo} presents a hierarchical audio-driven visual synthesis approach with the bounding box masks of lip, expression and head, it attains refined control over the diversity of expressions and variations in pose. 

\section{Method}

LinguaLinker aims to achieve realistic zero-shot audio-driven talking head generation. Given a reference portrait image and input audio, our approach ensures the preservation of portrait identity and faithful facial expressions, all while maintaining a harmonious alignment with the lip synchronization of the provided vocal audio. We present an overview of the framework in Fig \ref{fig:arch} and provide further details below.

\subsection{Architecture Design}
\label{sec:arch}

\textbf{Audio Encoder.} Some of the previous works~\cite{wei2024aniportrait, wang2024V-Express, xu2024hallo} employ Wav2Vec2-960h as the audio encoder, but in order to support multilingual audio inputs, we leverage a more powerful model, Wav2Vec2-XLS-R~\cite{babu2021xlsrselfsupervisedcrosslingualspeech}, which is claimed to be pre-trained on 436k hours of unlabeled speech in 128 languages. In our experiments, we found that Wav2Vec2-XLS-R  also responds well to languages other than English. We believe that its rich multilingual pretraining audio data enables it to distinguish subtle differences in various multilingual audio signals. After the feature extraction, the features from all Wav2Vec2 transformer blocks are averaged. We insert an MLP module between the audio encoder and the denoising net, which facilitates the projection of features from the audio feature space to the denoising net feature space. Before applying the cross-attention process on latents and projected audio embeddings, we transform the sequence of audio embedding to several blocks, from shape $(b \times f \times c)$ to $(b \times F \times l \times c)$, where $f$ represents original frames, $F$ indicates the generated video length and $l$ denotes the audio context length of the current frame. Each block represents the audio information corresponding to a specific video frame. This audio transformation operation enhances our model's ability to capture the audio information of the current frame, thereby enabling more accurate reconstruction of lip movements in the portrait.

\textbf{Reference Net.} We follow previous works~\cite{hu2023animateanyone}, applying ReferenceNet, which has been proven to be an effective framework design for extracting features of reference images. As it has the same architecture as the denoising net, it can be conveniently integrated with the generation pipeline. Due to the different dimensions of the features from the reference net, representing the reference features from high-level to low-level, the pipeline is capable of achieving high-fidelity results in the image or video generation process. 

\textbf{Denoising Net.} As shown in Fig \ref{fig:arch}, we modified the denoising net from the original stable diffusion network. To process reference images and audio signals, we enhance the cross-attention module to allow it to simultaneously receive two different types of conditions. Additionally, we introduce a region-specific gate mechanism~\cite{han2024emma} in this module, which is calculated by region masks and the corresponding gate offsets based on the input audio and denoising timestep. We compute the output results through the following equation:
\begin{equation}
    out = hidden\_states_{ref} + RegionGate * scale * hidden\_states_{audio}
    \label{equ1}
\end{equation}
Here, $scale$ is a weight factor and the model reconstructs the portrait randomly if $scale = 0$. The $RegionGate$ is derived from 
\begin{equation}
    RegionGate = RegionMask + \texttt{Tanh}(\texttt{Linear}(gate\_states)) * \texttt{SiLU}(alpha)
    \label{equ2}
\end{equation}
where $alpha$ is a learnable scale and the $gate\_states$ follows the calculation below:
\begin{equation}
    gate\_states = \texttt{AdaLN}(\texttt{Linear}(hidden\_states_{audio}), timestep)
\end{equation}
Besides, two types of $hidden\_states$ are:
\begin{equation}
    hidden\_states_{ref} = \texttt{CrossAttn}(hidden\_states, ref\_features)
\end{equation}
\begin{equation}
    hidden\_states_{audio} = \texttt{CrossAttn}(hidden\_states_{ref}, audio\_emb_{block})
\end{equation}
The $RegionGate$, $scale$, $alpha$ and $\texttt{Linear}$ layer in Equation\ref{equ2} are corresponding to three different regions $\{head, mouth, eyes\}$. With the region-specific gate, we can obtain the modification increment at the corresponding positions based on the audio information, building upon the original reference image. After the feature fusion, the temporal module~\cite{guo2023animatediff} is involved and contains self-attention in the time dimension. To ensure the consistency of video generation, similar to some previous works~\cite{tian2024emo, xu2024hallo}, we also include preceding frames as auxiliary information in the computation of temporal self-attention. By employing this mechanism, our model can generate more stable and coherent results, which are also better matched with the emotions and sentiments expressed in the audio.

\begin{figure}
  \centering
  \includegraphics[width=1\linewidth]{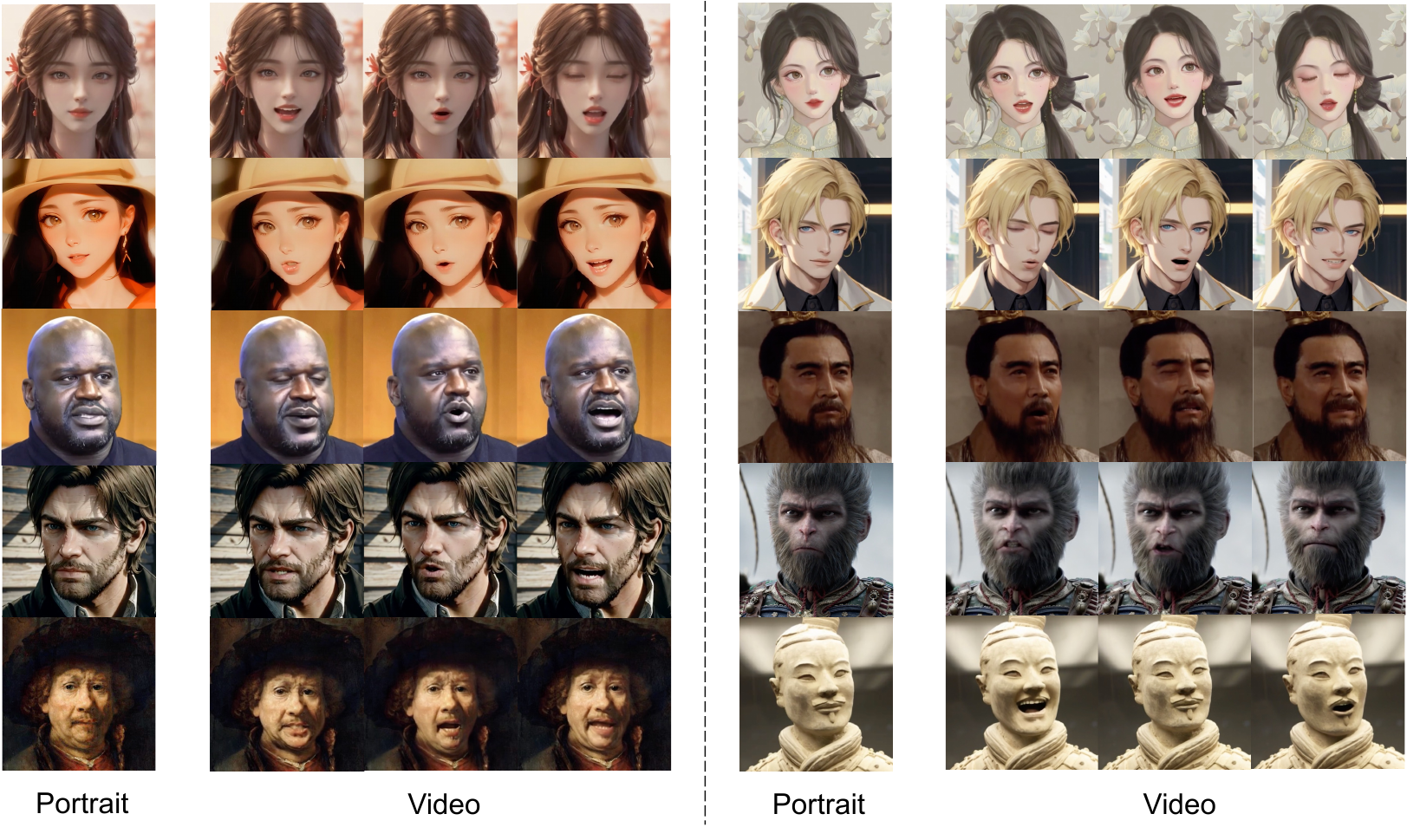}
  \caption{More generated results of diverse reference portraits and different languages.}
  \label{fig:more_results}
\end{figure}

\subsection{Training and Inference}

\label{sec:data}
\begin{figure}
  \centering
  \includegraphics[width=0.9\linewidth]{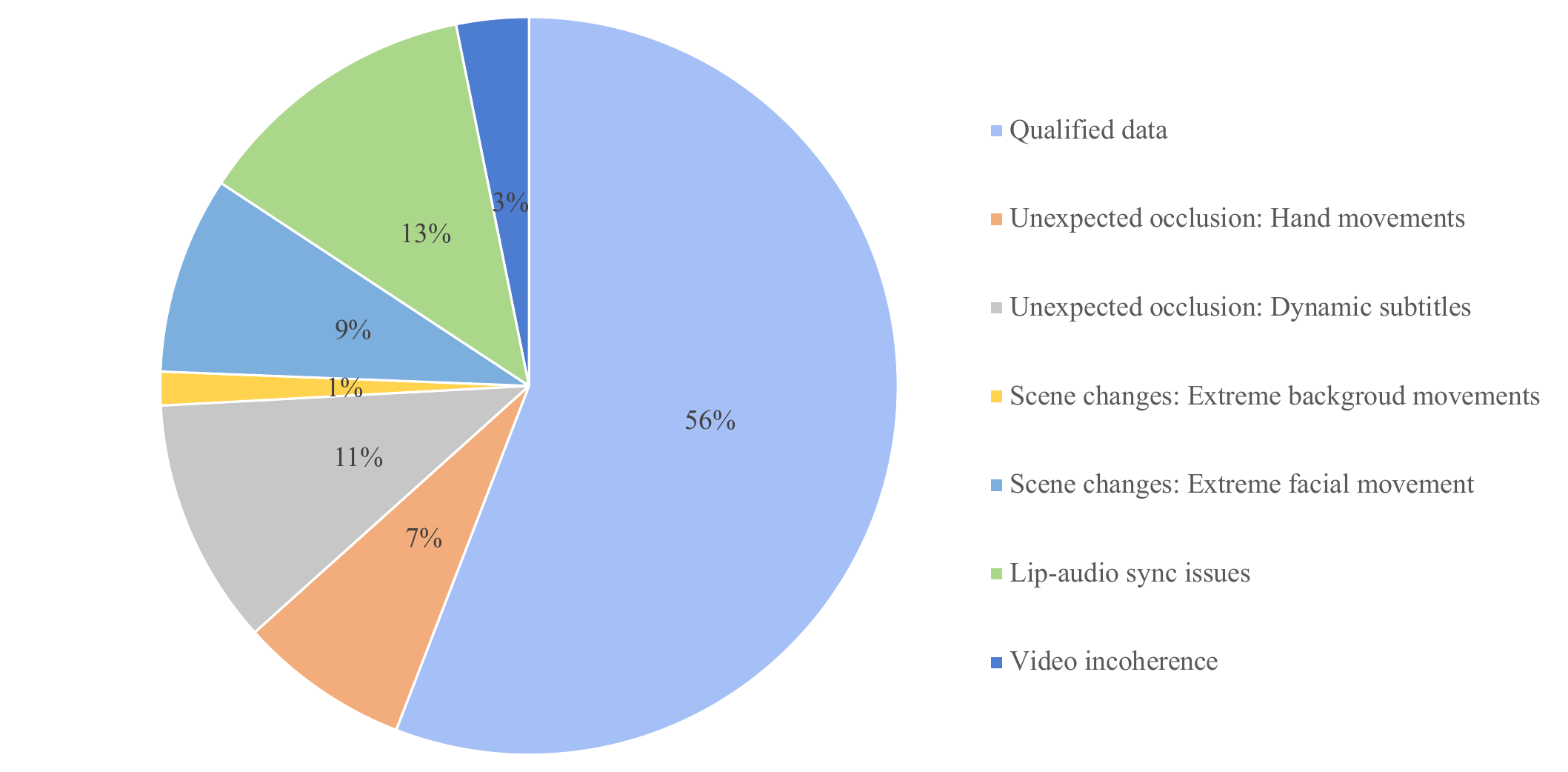}
  \caption{Web-sourced video data statistics.}
  \label{fig:data}
\end{figure}

\textbf{Data Pipeline.} It is well known that, as mentioned in \cite{blattmann2023stablevideodiffusionscaling}, training with a large amount of high-quality data helps to promote the performances of the models. Among the publicly accessible video datasets, HDTF~\cite{zhang2021flow} dataset has been a popular choice despite being limited to English-language speech content. To augment this, we have additionally gathered around 215 hours of talking head videos from the web to train our models. However, if the data cleaning process is not thorough, unqualified data may be mixed into the final training dataset, bringing some generation artifacts in the results. For example, we've observed that excessive object motion or drastic background changes in training videos can cause instability in the generated video outcomes, or lead to fluctuating video brightness over time. Moreover, certain splice points within the training video clips can also introduce similar issues, thereby affecting the coherence. We conducted a manual sampling of the data collected from the internet and found that it indeed contained the aforementioned issues and others. We have compiled and analyzed the results of this manual sampling, as depicted in Fig \ref{fig:data}. To enhance the quality of our data for model training, we meticulously process the collected footage through a four-stage filtration process: (1) Unexpected occlusion issue; (2) Extreme scene changes; (3) Misalignment of lip movements and audio signals; (4) Incoherence of the video. After this rigorous curation, we've refined our dataset from an initial 215 hours down to approximately 114 hours (Tab \ref{tab:data-statistics}), ensuring it's primed for effective model training. 

\begin{table}[h]
  \centering
  \begin{tabular}{llllll}
    \toprule
    Curation Stage & Origin & Stage 1 & Stage 2 & Stage 3 & Stage 4 \\
    \midrule
    Data Size (hours) & 215 & 156 & 153 & 121 & 114 \\
    \bottomrule
  \end{tabular}
  \vspace{10pt}
  \caption{Data statistics of curation pipeline.}
  \label{tab:data-statistics}
\end{table}
% \vspace{-10pt}

\textbf{Loss Design.} During the training process, we observed that in some cases, the generated portrait videos exhibited unexpected lip movements and the eyes appeared rigid and unnatural. To tackle these issues, we add additional region loss for the mouth and eyes(representing the eyes and eyebrows region in this paper) and adjust the weights of the loss function, keeping the balance of the overall loss and regional loss weights. This adjustment helps the model to pay more attention to the compatibility of facial movements and input audio and makes the generated portraits more vivid and natural.

\textbf{Inference.} As we implicitly split the control signal to head, mouth and eyes, we could adjust the corresponding parameters to influence the generated portraits and separately control the motion range. We note that the $scale$ in Equation\ref{equ1} facilitates the integration of audio content into visual features, thereby enabling the modulation of motion frequencies across various regions associated with the character. Moreover, to stabilize the background of the long-term video generation, especially around 60 seconds in length, we apply the image-to-image generation pipeline for the background part, which could reduce the artifacts and error accumulation beyond the portraits.

\section{Experiment}

\subsection{Inplementation Details}

The training consists of two stages. The first stage is image pertaining, aiming to endow the model with the ability to reconstruct the input portrait image while maintaining the diversity of the generated outputs, such as different lip movements and head poses. The second stage involves training for video generation, incorporating the temporal modules and an audio encoder, which enables the model to generate multi-frame, lip-synchronized videos. In the second stage, the expected video length of the output is 12 frames and 4 preceding frames are employed for video coherence. The audio context frame equals 5, which means the current frame generation utilizes audio information from the preceding and succeeding 2 frames additionally. For both stages, the images or video clips sampled from the dataset are resized and cropped to 512 × 512 and the learning rate is set to 1e-5.

\begin{figure}
  \centering
  \includegraphics[width=0.55\linewidth]{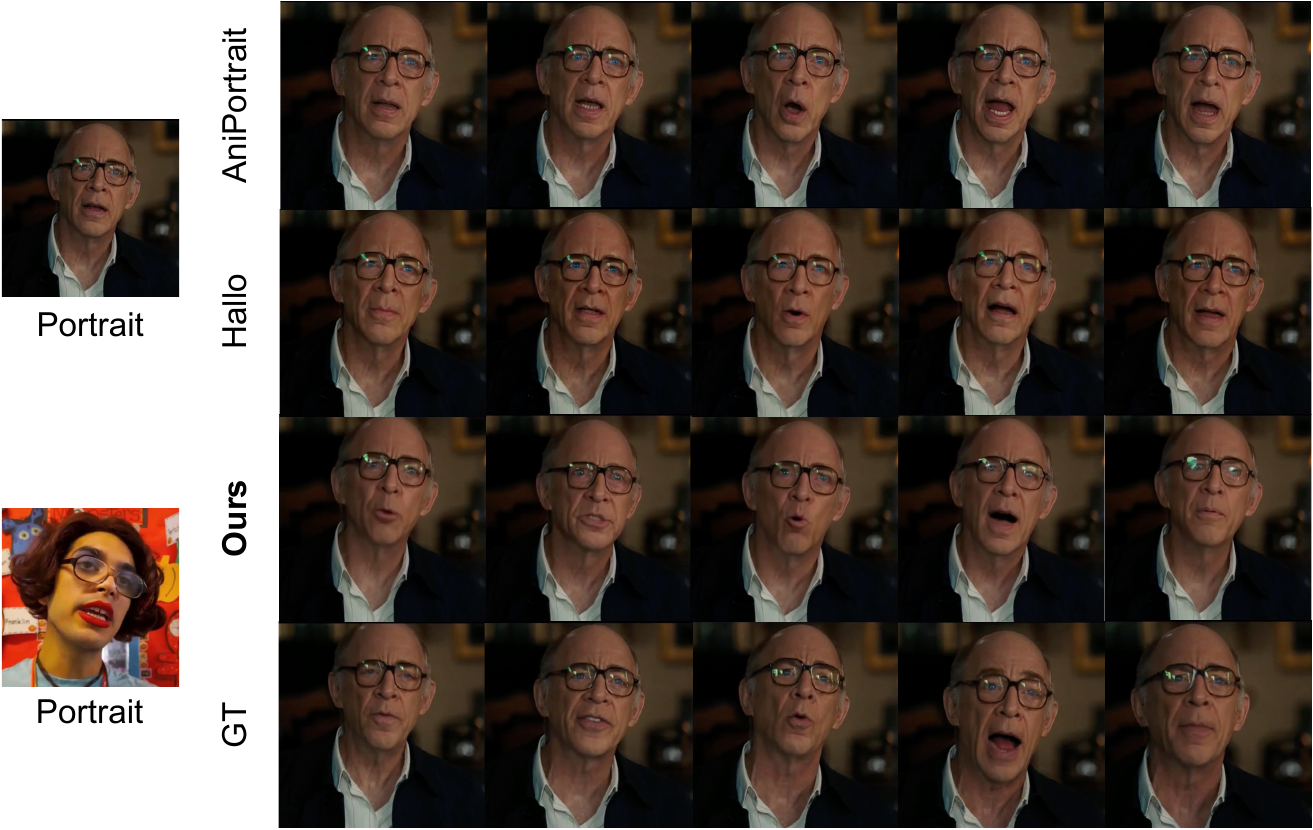}
  \includegraphics[width=0.432\linewidth]{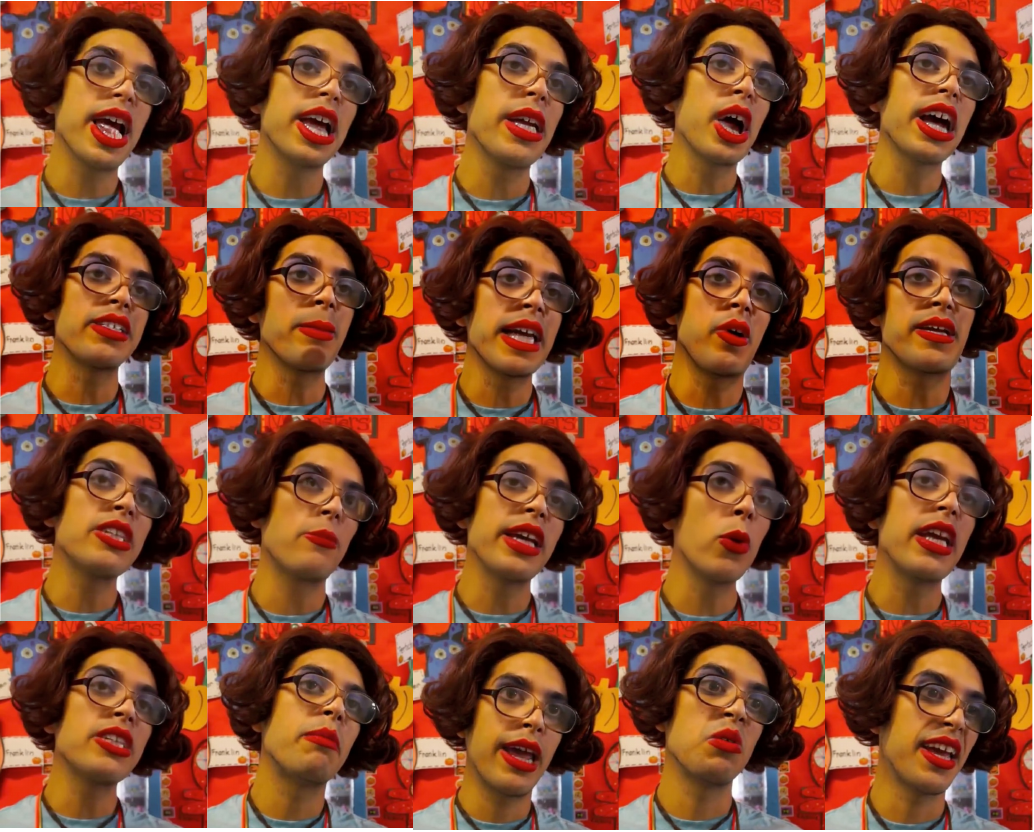}
  \caption{The comparison between different methods and the ground truth videos are sampled from CelebV-HQ dataset.}
  \label{fig:method_comparison}
\end{figure}

\subsection{Evaluation}

To better evaluate our model, we adopt 4 evaluation metrics, which are FID, FVD, SSIM, and Sync. FID (Fréchet Inception Distance) is a widely used metric for evaluating the quality of generated images. It assesses the performance of generative models by calculating the distance between the generated images and real images in the feature space. FVD (Fréchet Video Distance) extends the concept of FID to the video domain. SSIM (Structural Similarity Index Measurement) is a metric used to evaluate image quality. It measures the similarity between two images based on their structural information. Sync here is a metric based on SyncNet~\cite{Chung16a} to calculate the audio-visual synchronization for each frame of the person in the video. It provides the average confidence score, indicating how well the audio and visual components are synchronized.

Tab \ref{tab:hdtf-eval} and Tab \ref{tab:celebv-eval} show our quantitative results on HDTF~\cite{zhang2021flow} and CelebV-HQ~\cite{zhu2022celebvhq} compared with other methods and the \textbf{bold} score denotes the best performance. For both datasets, we filtered and sampled around 1000 video clips as the test set. To provide more intuitive comparison results, a visual comparative examination is presented in Fig \ref{fig:method_comparison}.

\begin{table}[h]
  \centering
  \begin{tabular}{l|llll}
    \toprule
    Methods & FID $\downarrow$ & FVD $\downarrow$ & SSIM $\uparrow$ & Sync $\uparrow$ \\
    \midrule
    AniPortrait & 9.301 & 2296.930 & 0.717 & 2.380 \\
    Hallo & \textbf{5.275} & 2454.547 & \textbf{0.755} & 4.262 \\
    \textbf{LinguaLinker (Ours)} & 5.284 & \textbf{1156.387} & 0.745 & \textbf{4.770} \\
    \midrule
    Real Video & - & - & - & 4.292 \\
    \bottomrule
  \end{tabular}
  \vspace{10pt}
  \caption{The quantitative results on HDTF test dataset.}
  \label{tab:hdtf-eval}
\end{table}
\vspace{-5pt}

\begin{table}[h]
  \centering
  \begin{tabular}{l|llll}
    \toprule
    Methods & FID $\downarrow$ & FVD $\downarrow$ & SSIM $\uparrow$ & Sync $\uparrow$ \\
    \midrule
    AniPortrait & 14.021 & 2322.836 & 0.589 & 2.886 \\
    Hallo & 11.119 & 1787.557 & 0.594 & 6.024 \\
    \textbf{LinguaLinker (Ours)} & \textbf{9.921} & \textbf{897.912} & \textbf{0.598} & \textbf{6.132} \\
    \midrule
    Real Video & - & - & - & 6.886 \\
    \bottomrule
  \end{tabular}
  \vspace{10pt}
  \caption{The quantitative results on CelebV-HQ test dataset.}
  \label{tab:celebv-eval}
\end{table}

\subsection{Results Comparison and Analysis}

\begin{figure}
  \centering
  \includegraphics[width=1\linewidth]{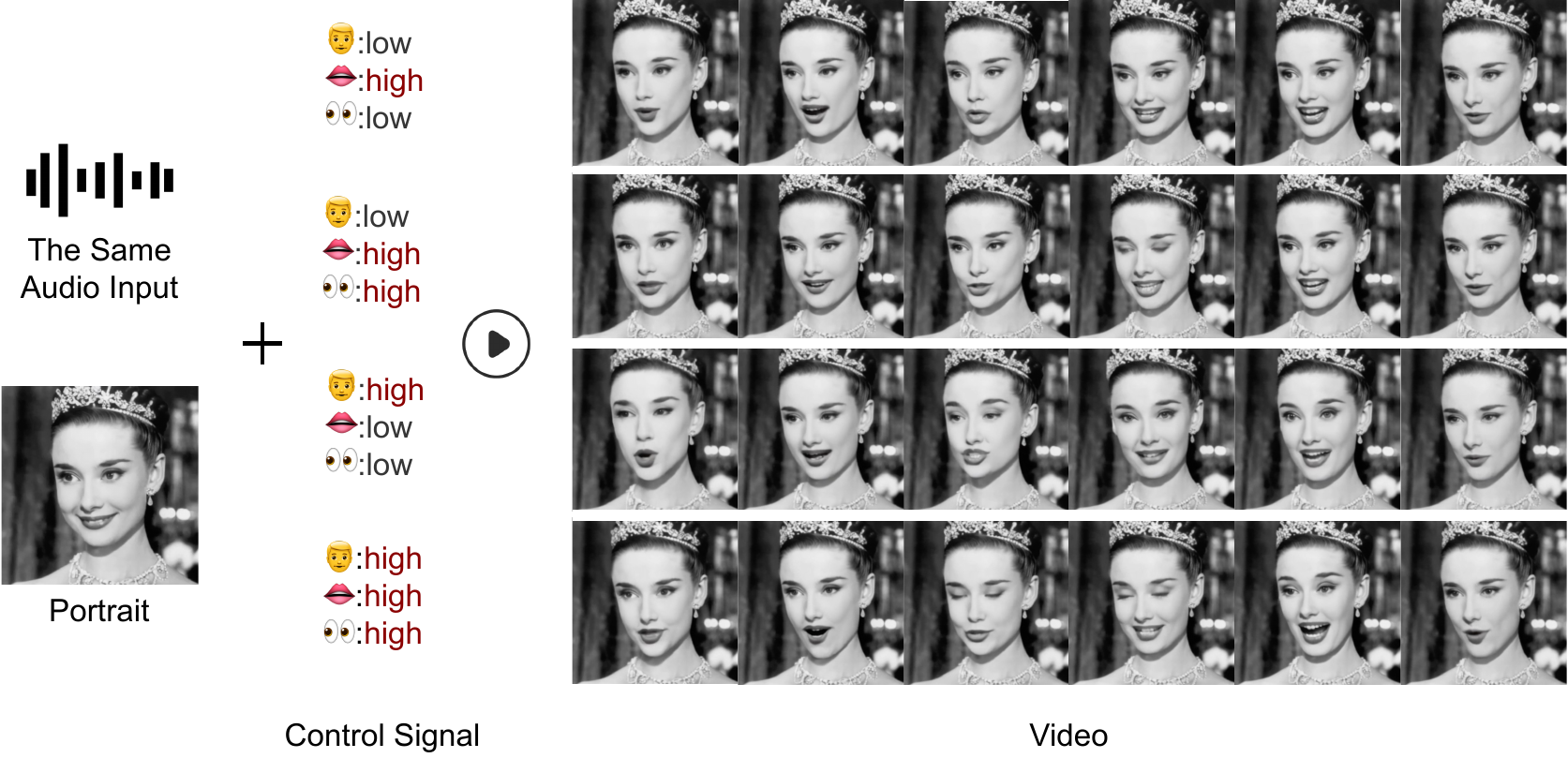}
  \caption{The comparison of different implicit control signals.}
  \label{fig:comparison}
\end{figure}

Fig \ref{fig:more_results} shows the results generated with diverse given portraits and audios in different languages. The types of portraits include AI-generated personas, 2D/3D animated characters, realistic people, sculptures, oil paintings, etc. The input portraits could speak Chinese, English, Japanese, Korean, French, and Thai in the output videos.

During training, the area of the region mask is determined by the motion range of the corresponding target. For instance, an increase in the mask area of the head indicates a larger range of head movement, while an increase in the mask area of the eyes may suggest pronounced eyebrow movements in the video. Thus we could simply adjust the area of the region mask to influence the motion ranges of the head, eyes and mouth during inference. Besides, the $scale$ can be called the frequency parameter, which is responsible for frequency control. Some comparison results are shown in Fig \ref{fig:comparison}. Comparing the first row and the second shows that the lower eye frequency parameter means the character in the video almost stops blinking. Additionally, compare the results of the first two rows with those of the last two rows, the lower head frequency parameter generates a speaker with fewer head movements in the video. 

From our results comparison and analysis, the control parameters provided by our method can slightly adjust the range of motion and frequency of actions of the characters in the generated video, without affecting the overall expression of the audio's emotions, which ensures the consistency and compatibility between the output videos and the input audios.

\section{Conclusion}

In this paper, we introduce LinguaLinker, an end-to-end audio-driven portrait animation approach based on diffusion techniques, which is also supported to implicitly control the portrait motion range and frequency during the inference stage through the design of the region-specific gates. LinguaLinker is capable of generating videos with any type of portrait image and audio in multiple languages as inputs. The generated results demonstrate lip-synchronization, compatibility with audio, high fidelity and coherence. But in terms of limitations, the inference of process is time-consuming, a common issue with diffusion models. Additionally, although LinguaLinker supports multilingual audio signals, the lip synchronization performance varies slightly across different languages. The results also exhibit some generation artifacts, particularly in fine textures and embellishments, or blurred frames, which can affect coherence. In the future, we are committed to addressing these issues to achieve more refined generation results.

\newpage
\bibliographystyle{unsrtnat}
\bibliography{neurips_2024}

\end{document}